\documentclass{article}

\usepackage{PRIMEarxiv}

\usepackage[utf8]{inputenc} 
\usepackage[T1]{fontenc}    
\usepackage{hyperref}       
\usepackage{url}            
\usepackage{booktabs}       
\usepackage{amsfonts}       
\usepackage{nicefrac}       
\usepackage{microtype}      
\usepackage{lipsum}
\usepackage{fancyhdr}       
\usepackage{graphicx}       
\graphicspath{{media/}}     
\usepackage{amssymb}
\usepackage{amsmath}
\usepackage{multirow}
\usepackage{multicol}
\usepackage{graphicx,subcaption}
\title{PU-MFA : Point Cloud Up-sampling via Multi-scale Features Attention}

\author{
  Hyungjun Lee\\
  Graduate School of Automotive Engineering \\
  Kookmin University \\
  Seoul\\
  \texttt{rhtm13@kookmin.ac.kr} \\
   \And
   Sejoon Lim  \\
  Department of Automobile and IT Convergence \\
  Kookmin University \\
  Seoul\\
  \texttt{lim@kookmin.ac.kr} \\
}

\begin{document}
\maketitle

\begin{abstract}
\quad Recently, research using point clouds has been increasing with the development of 3D scanner technology. According to this trend, the demand for high-quality point clouds is increasing, but there is still a problem with the high cost of obtaining high-quality point clouds.  Therefore, with the recent remarkable development of deep learning, point cloud up-sampling research, which uses deep learning to generate high-quality point clouds from low-quality point clouds, is one of the fields attracting considerable attention. This paper proposes a new point cloud up-sampling method called Point cloud Up-sampling via Multi-scale Features Attention (PU-MFA). Inspired by previous studies that reported good performance using the multi-scale features or attention mechanisms, PU-MFA merges the two through a U-Net structure. In addition, PU-MFA adaptively uses multi-scale features to refine the global features effectively. The performance of PU-MFA was compared with other state-of-the-art methods through various experiments using the PU-GAN dataset, which is a synthetic point cloud dataset, and the KITTI dataset, which is the real-scanned point cloud dataset. In various experimental results, PU-MFA showed superior performance in quantitative and qualitative evaluation compared to other state-of-the-art methods, proving the effectiveness of the proposed method. The attention map of PU-MFA was also visualized to show the effect of multi-scale features.
\end{abstract}


\section{Introduction}
\label{sec:introduction}
\quad A point cloud is one of the most popular formats for accurately representing 3D geometric information in robotics and autonomous vehicles. Recently, the number of studies using point clouds has been increasing with the development of 3D scanners, such as LiDAR \cite{koide2019portable,lim2022single}. Along with this trend, there is an increasing demand for high-quality point clouds that are low-noise, uniform, and dense. However, the high cost of collecting high-quality point clouds remains problematic. Therefore, point cloud up-sampling, which generates a low-noise, uniform, and dense point set from noisy, non-uniform, and sparse point sets, is an interesting study.

\quad Similar to learning-based image super-resolution studies \cite{niu2020single,liang2021swinir}, various learning-based point cloud up-sampling studies \cite{qiu2021pu,luo2021pu,qian2021pu} show better performance than traditional point cloud up-sampling studies \cite{alexa2003computing,lipman2007parameterization}. Intuitively, the image super-resolution tasks and the point cloud up-sampling tasks are similar. Unlike the image super-resolution tasks, which process regular format images, the point cloud up-sampling tasks, which process irregular formats, require additional consideration. First, the up-sampled point set should have a uniform distribution and a dense set of points. Next, the up-sampled point set should represent the details of the target 3D mesh surface well \cite{li2019pu}.

\quad A traditional learning-based point cloud up-sampling study usually consists of a feature extractor and an up-sampler. In addition, most studies use multi-scale features or attention mechanisms. PU-Net \cite{yu2018pu}, 3PU \cite{yifan2019patch}, and PU-GCN \cite{qian2021pu} extract multi-scale features from sparse point sets. These studies reported excellent performance, but the last feature extracted by the feature extractor has a limitation in that the details of the sparse point set are diluted features because the output of each layer is used as the input of the next layer. Dis-PU \cite{li2021point}, PU-EVA \cite{luo2021pu}, and PU-Transformer \cite{qiu2021pu} showed successful performance using the self-attention mechanism to learn long-range dependencies between points. However, there is a limit to applying the attention mechanism with limited information because the key, query, and value of the self-attention mechanism are generated from the same input.

\quad Focusing on these limitations, this paper proposes PU-MFA, a novel method to fuse multi-scale features and attention mechanisms. PU-MFA solves point cloud up-sampling through an attention mechanism that uses an adaptive feature for each layer. The contributions of this research are as follows:
\begin{itemize}
    \item This paper proposes a point cloud up-sampling method of U-Net structure using Multi-scale Features (MFs) adaptively to Global Features (GFs).

    \item Global Context Refining Attention (GCRA), a structure for effectively combining MFs and attention mechanisms, is proposed. To the best of the authors’ knowledge, this is the first MultiHead Cross-Attention (MCA) mechanism proposed in point cloud up-sampling.
    
    \item This study demonstrates the effect of MFs by visualizing the attention map of GCRA in ablation studies.
\end{itemize}
This method was compared with various state-of-the-art methods using the Chamfer Distance (CD), Hausdorff Distance (HD), and Point-to-Surface (P2F) evaluation metrics for the PU-GAN \cite{li2019pu} and the KITTI \cite{geiger2013vision} dataset. As a result, the effectiveness of this method was confirmed by showing better performance.

\section{Related Work}
\label{sec:related_works}
\subsection{Optimization-Based Point Cloud Up-Sampling}
\quad Various optimization-based studies have been performed to generate a dense set of points from a sparse set. Alexa \textit{et al.} solved up-sampling by inserting new points into the Voronoi diagram of the local tangential space computed based on the moving-least-squares error \cite{alexa2003computing}. Lipman \textit{et al.} explained the up-sampling using the Locally Optimal Projection (LOP) operator \cite{lipman2007parameterization}. In this study, the points were re-sampled by using $L_1$ norm. Huang \textit{et al.} up-sampled a noisy and non-uniform set of points using an improved LOP that is a weighted LOP \cite{huang2009consolidation}. Later, Huang \textit{et al.} proposed an advanced method called Edge-Aware Re-sampling of a set of points (EAR). The EAR first re-samples the edges and then uses edge-aware up-sampling to resolve the up-sampling \cite{huang2013edge}.

\subsection{Learning-Based Point Cloud Up-Sampling}
\quad With the successful performance of learning-based image super-resolution, many studies have proposed a learning-based point cloud up-sampling method.

\quad As with image analysis, many studies have used MFs in point cloud up-sampling. PU-Net \cite{yu2018pu}, the first attempt at deep learning for point cloud up-sampling, showed good performance by extracting MFs through hierarchical feature learning and interpolation based on the framework of PointNet++ \cite{qi2017pointnet++}. 3PU \cite{yifan2019patch} performed well using MFs via an Intra-Level Dense connection and Inter-Level Skip connection. PU-GCN \cite{qian2021pu} uses MFs extracted by Inception DenseGCN. In this study, Inception DenseGCN could effectively extract MFs with an InceptionNet-inspired structure \cite{szegedy2015going}.

\quad Because of the advantages of learning the long-range dependency of the self-attention mechanism, it is used in various point cloud up-sampling studies. In PU-GAN \cite{li2019pu}, generators are trained using discriminators that apply a self-attention mechanism. Pugeo-Net \cite{qian2020pugeo} showed good performance by using it in Feature Recalibration. PU-EVA \cite{luo2021pu} showed successful up-sampling performance using an EVA Expansion Unit with the mechanism. Dis-PU \cite{li2021point} performed well using the Local Refinement Unit with self-attention applied to the generated point set. PU-Transformer \cite{qiu2021pu}, which applied the transformer structure for the first time in point cloud up-sampling, uses Shifted Channel MultiHead Self-Attention to show the state-of-the-art performance.

\section{Problem Description}
\label{sec:problem_description}
\begin{figure}[htb!]
  \centering  \includegraphics[width=\columnwidth]{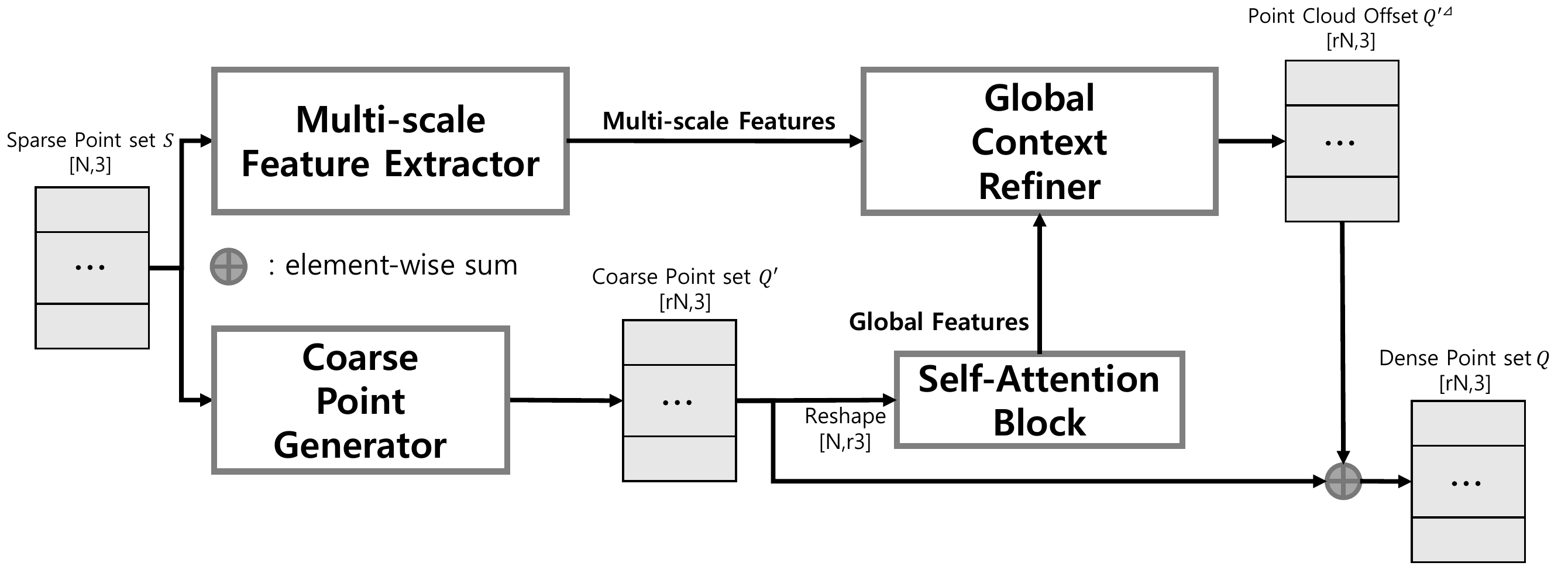}
  \caption{Illustration of an overview of the proposed method.}
  \label{fig:overview}
\end{figure}

\quad The problem of generating low-noise, uniform, and dense point set $Q = \left\{q_{i} \right\}_{i = 1}^{rN}$ was addressed using the Ground Truth (GT) point set $D = \left\{d_{i} \right\}_{i = 1}^{rN}$ and sparse point set $S = \left\{s_{i} \right\}_{i = 1}^{N}$, where $N$ is the input patch size and $r$ is the up-sampling ratio. Figure \ref{fig:overview} shows the problem description of this study.

\section{Method}
\label{sec:method}
\begin{figure*}[htb!]
  \centering
  \includegraphics[width=\textwidth]{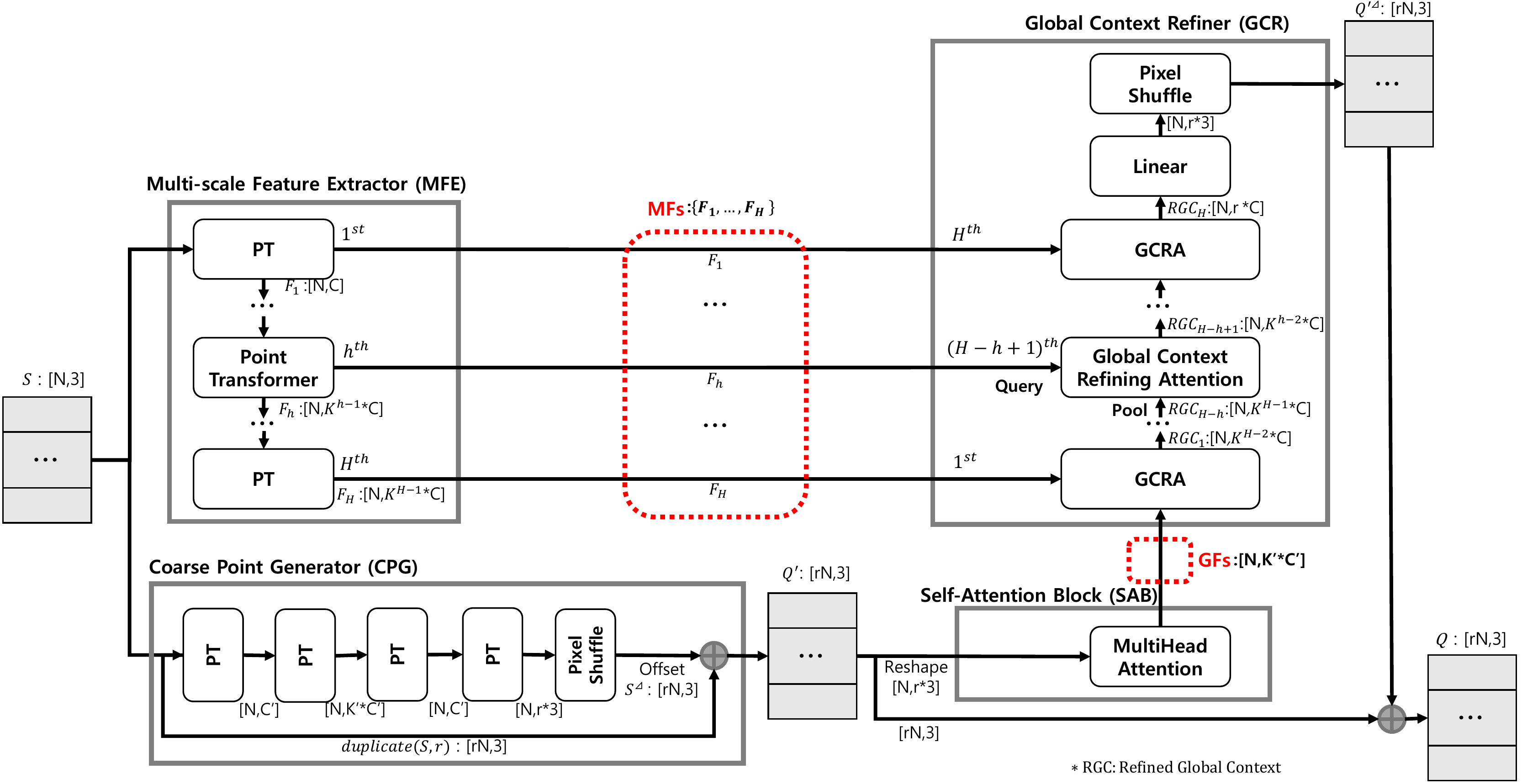}
  \caption{Illustration of the proposed framework. Here, 3 is the coordinate dimension, and $H$ is the depth of the layer. In Multi-scale Feature Extractor (MFE) and Global Context Refiner (GCR), $C$ is the channel, and $K$ is the expansion ratio. In Coarse Point Generator (CPG) and Self-Attention Block (SAB), $C’$ is the channel and $K'$ is the expansion ratio.}
  \label{fig:model}
\end{figure*}

\quad This method consists of a Multi-scale Feature Extractor (MFE), Global Context Refiner (GCR), Coarse Point Generator (CPG), and Self-Attention Block (SAB). As shown in Figure \ref{fig:model}, MFE extracts MFs, an adaptive feature for use in GCR. GCR uses MFs to refine GFs adaptively and finally produce $Q'^{\Delta}$, where $Q'^{\Delta}$ is defined as $\mathit{Q'^{\Delta}} = \left\{\mathit{q'^{\Delta}}_{i} \right\}_{i = 1}^{rN}$. CPG generates $Q'$ from $S$ and SAB extracts GFs from $Q'$, where $Q'$ is defined as $\mathit{Q'} = \left\{\mathit{q'}_{i} \right\}_{i = 1}^{rN}$. Based on the definitions of $Q'$ and $\mathit{Q'^{\Delta}}$, $Q$ is formulated as equation (\ref{refining}), where $\oplus$ is an element-wise sum.
\begin{center}
\begin{equation}\label{refining}
\scalebox{0.87}{%
$\begin{aligned}
    Q=Q'\oplus Q'^{\Delta} \left (\oplus:\textit{element-wise sum}\right )
\end{aligned}$%
}
\end{equation}
\end{center}

\subsection{Multi-scale Feature Extractor}
\label{sec:cascade_local_feature_extractor}
\quad Because the GFs extracted from $Q'$ via SAB is a feature extracted from a set of points in which geometric information about the original input $S$ is diluted, MFE using Point Transformer (PT) \cite{zhao2021point}, an advanced point cloud analysis technique, extracts MFs from $S$. As shown in Figure \ref{fig:model}, the MFE consists of $H$ PT, and the set of point-wise features extracted from the $h^{th}$ PT is $F_h \in \mathbb{R}^{N \times K^{h-1}C}$. MFs are the set of $F_h$ extracted from all layers of the MFE. The extracted MFs and $F_h$ are formulated as in equation (\ref{clfs}), where $f_i^h$ is a point-wise feature extracted from the $h^{th}$ PT.

\begin{center}
\begin{equation}\label{clfs}
\scalebox{0.87}{%
$\begin{aligned}
    &F_h=\left\{f_{i}^{h}\right\}_{i=1}^N, MFs=\left\{F_h\right\}_{h=1}^H
\end{aligned}$%
}
\end{equation}
\end{center}

\subsubsection{Point Transformer} 
\begin{figure}[htb!]
  \centering
 \subfloat[$h^{th}$ Point Transformer layer with $s_i$]{\includegraphics[width=0.5\columnwidth]{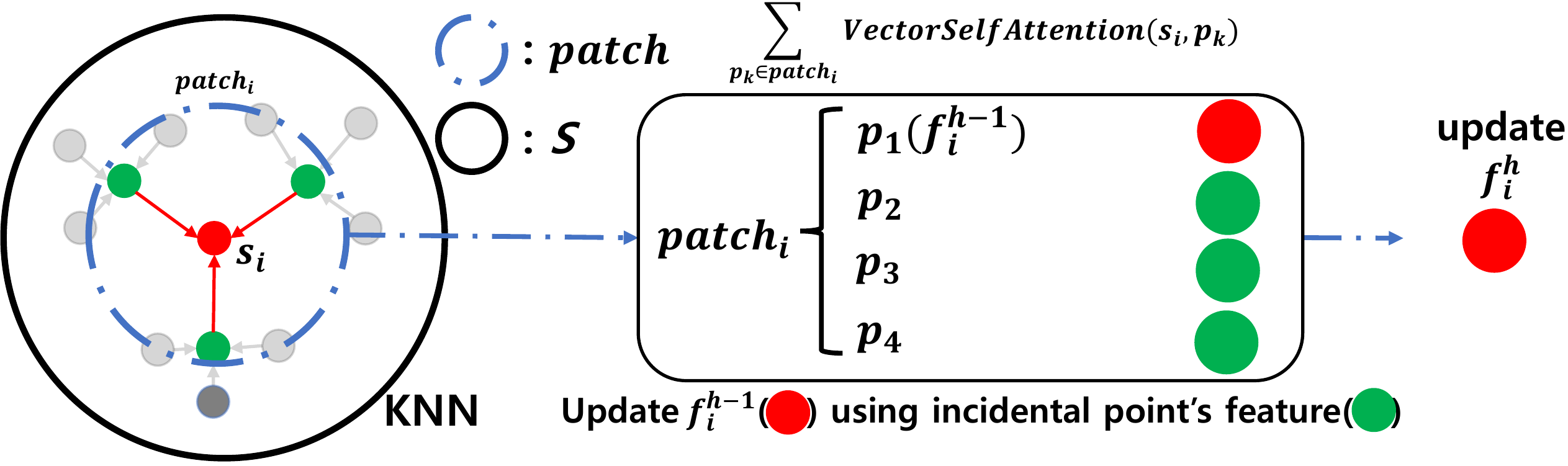}}
 \subfloat[$h^{th}$ Point Transformer layer with incidental points]{\includegraphics[width=0.5\columnwidth]{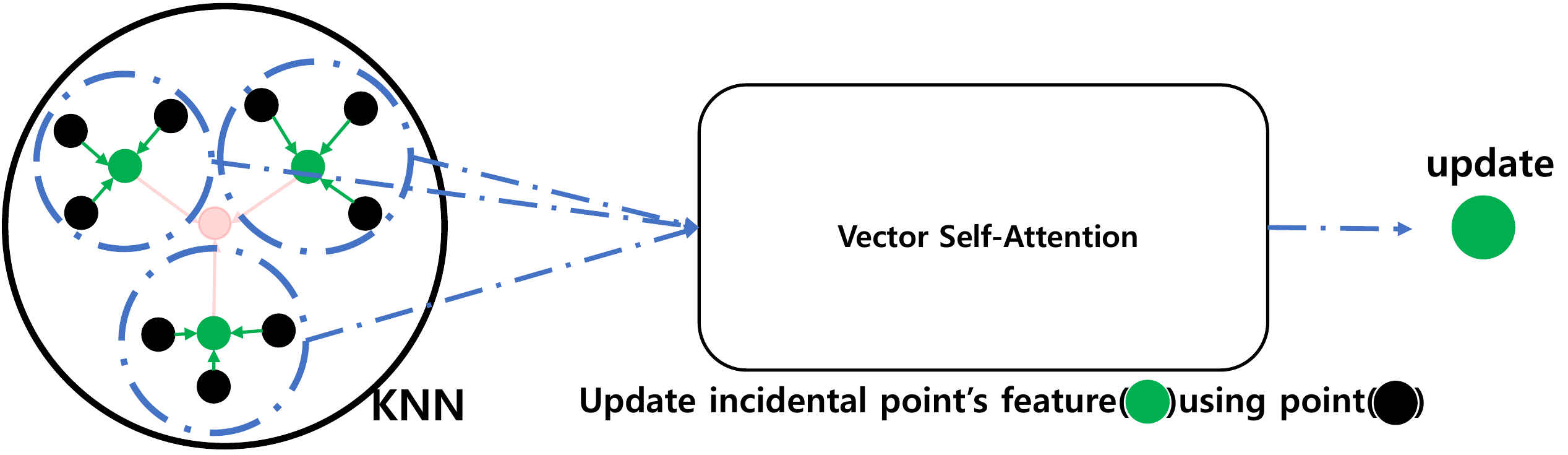}} \quad
 \subfloat[$(h+1)^{th}$ Point Transformer layer with $s_i$]{\includegraphics[width=0.5\columnwidth]{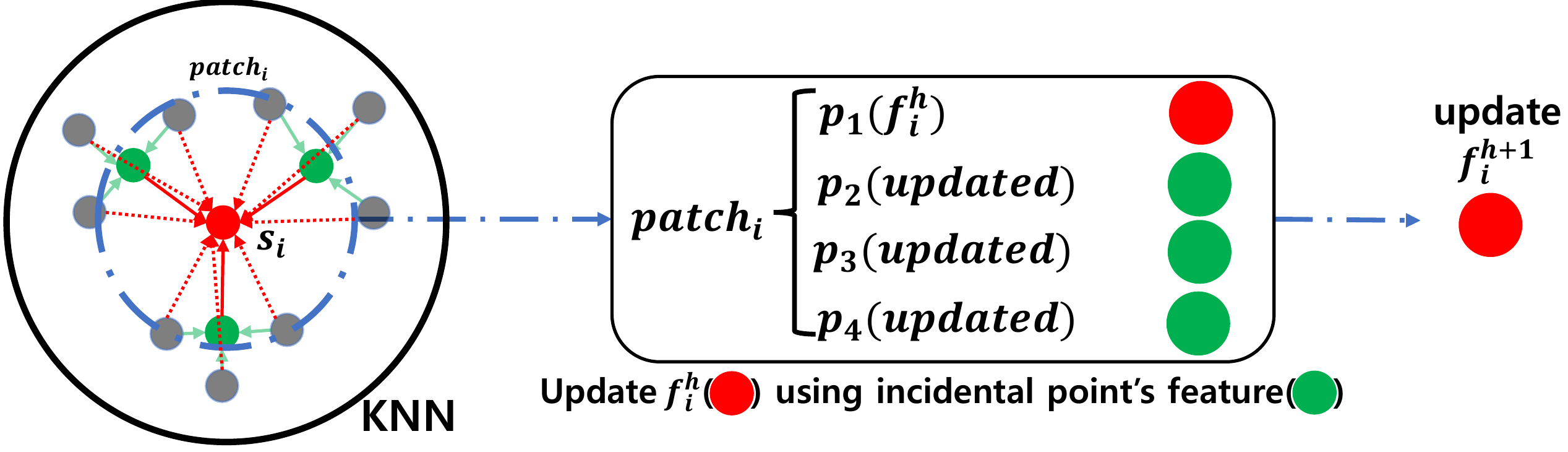}}
\caption{Illustration of KNN and VSA in PT.}
\label{fig:knn_operation}
\end{figure}

\quad PT consists of two elements. The first is the K-Nearest Neighbor (KNN), and the second is the Vector Self-Attention (VSA) mechanism. At the $h^{th}$ PT, the point-wise feature $f_i^{h-1}$ of point $s_i\in S$ is updated to $f_i^{h}$ through VSA, which uses $s_i$ and $patch_i$ as the inputs. The $patch_i$ is generated through KNN using $s_i$ as the input. This operation works on all points in $S$, updating the point-wise feature of all points \cite{zhao2021point}. This is formulated in equation (\ref{point_transformer}), where $patch\_size$ is the size of K in KNN.
\begin{center}
\begin{equation}\label{point_transformer}
\scalebox{0.87}{%
$\begin{aligned}
    &patch_i = KNN(s_i,patch\_size) \\
    &f_i^h = \sum_{p_k\in patch_i}VectorSelfAttention(s_i,p_k) \\
    &(s_i \in S, i\in \left\{1,2,...,N\right\})
\end{aligned}$%
}
\end{equation}
\end{center}
\quad Inspired by this operation, this considered $patch_i$ is equivalent to the CNN's kernel. In CNN, even if the kernel of the CNN is fixed, a deeper layer, means a wider receptive field. Therefore, even if the patch size of the KNN in PT is fixed, the deeper the layer, the more $s_i$ can interact with a wider range of points. Figure \ref{fig:knn_operation} is an example with a KNN patch size of four. In Figure \ref{fig:knn_operation} (a), when the $h^{th}$ PT updates $f_i^{h-1}$ to $f_i^{h}$, VSA is performed on $patch_i$, which is composed of $s_i$ and incidental points, to update $f_i^{h-1}$. In Figure \ref{fig:knn_operation} (b), the $h^{th}$ PT updates each feature by performing a VSA for each patch in all incidental points. In Figure \ref{fig:knn_operation} (c), the $(h+1)^{th}$ PT updates $f_i^h$ to $f_i^{h+1}$ by performing VSA using $patch_i$ similar to the $h^{th}$ PT. However, the $(h+1)^{th}$ PT updates $f_i^h$ to $f_i^{h+1}$ using a wider receptive field than the receptive field of the $h^{th}$ PT because the features of the incidental points of the $(h+1)^{th}$ PT are updated by the $h^{th}$ PT. This operation allows the MFE to extract the MFs effectively.

\subsection{Global Context Refiner}
\quad  Because GCR and MFE are U-Net \cite{ronneberger2015u} structures, they are composed of $H$ GCRA. GCRA effectively refines GFs by querying MFs, which is adaptive geometric information applied to each layer. As shown in Figure \ref{fig:model}, the $(H-h+1)^{th}$ GCRA  generates $RGC_{H-h+1}\in \mathbb{R}^{N \times K^{h-2}C}$ by using $F_h\in \mathbb{R}^{N \times K^{h-1}C}$ as a query and $RGC_{H-h}$ as a pool. However, $\mathbb{R}^{N \times r3}$ was used instead to prevent $RGC_{H}$ from becoming $\mathbb{R}^{N \times \frac{C}{K}}$. After refining the GFs, the linear layer was used to perform the transformation. PixelShuffle was then used to generate the $Q'^{\Delta}\in \mathbb{R}^{rN \times 3}$.

\subsubsection{Global Context Refining Attention} 
\begin{figure}[htb!]
  \centering
\includegraphics[width=0.7\columnwidth]{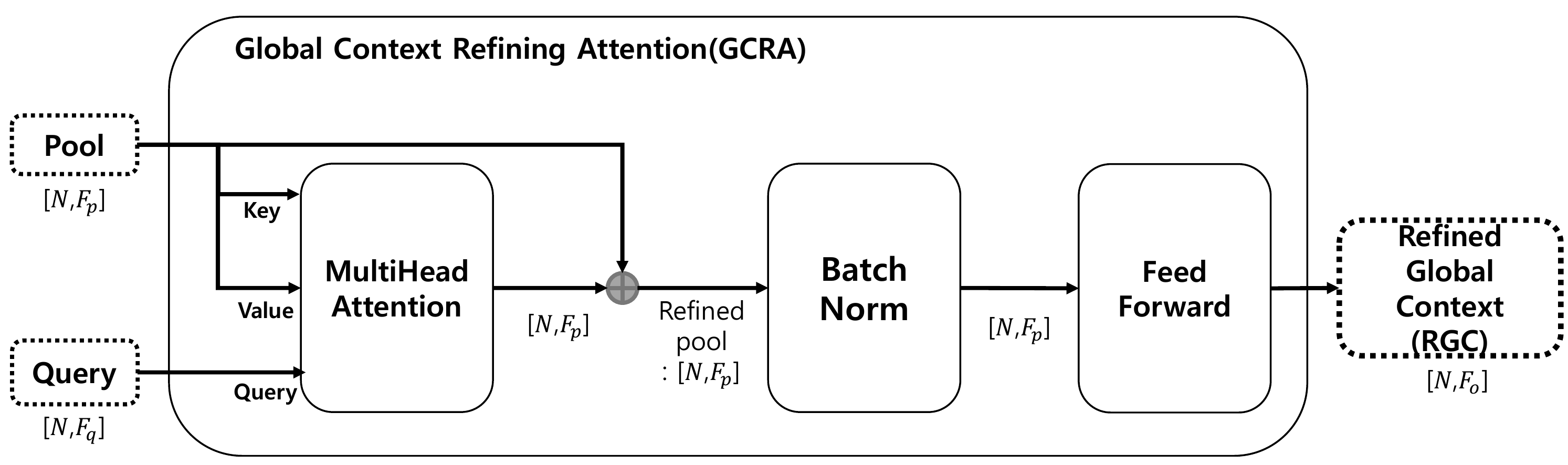}
\caption{Illustration of Global Context Refining Attention (GCRA). $F_p$ is the pool input channel, $F_q$ is the query input channel, and $F_o$ is the output channel.}
\label{fig:clf}
\end{figure}

\quad Inspired by Skip-Attention \cite{wen2020point}, which acts as a communicator between the encoder and decoder, GCRA uses MFs and GFs to apply the MCA mechanism. In various studies, self-attention mechanisms are used to extract the features of point sets or to generate up-sampling point sets \cite{qiu2021pu,li2019pu}. However, the self-attention mechanism is limited because it uses only limited information due to the structure in which key, query, and value are generated from the same input. With these limitations in mind, GCRA in the $H$ hierarchy uses GFs$\in \mathbb{R}^{N \times K'C'}$ as the pool (key, values) and MFs as the queries, progressively refining the GFs through MCA. GCRA consists of MCA \cite{vaswani2017attention}, Batch Normalization (BN) \cite{ioffe2015batch}, and Feed Forward. As shown in Figure \ref{fig:clf}, the output shape of applying MCA using the query and pool is $\mathbb{R}^{N \times F_p}$. The pool was then refined by adding the pool and the MCA output. The BN was used for stable training after addition. Feed Forward transforms the output of the BN and produces a Refined Global Context (RGC)$\in \mathbb{R}^{N \times F_o}$.

\subsection{Coarse Point Generator}
\label{sec:coarse_point_generator}
\quad CPG generates $Q'$. In CPG, PT \cite{zhao2021point} and PixelShuffle \cite{shi2016real,qiu2021pu} generate $S'^{\Delta}$ from $S$, where, $S'^{\Delta}$ is defined as $\mathit{S'^{\Delta}} = \left\{\mathit{s'^{\Delta}}_{i} \right\}_{i = 1}^{rN}$. The structure of CPG consists of four layers, such as the structure of the 3PU's Feature Extraction Unit \cite{yifan2019patch}. As shown in Figure \ref{fig:model}, to make the final output into 3D coordinates, first, PT was first used to expand the features, and then gradually reduce them. Subsequently, PixelShuffle generates 3D coordinates using those features. $Q'$ is generated through the element-wise sum of the generated $S'^{\Delta}$ and $duplicate(S,r)\in\mathbb{R}^{rN \times 3}$. This process is formulated as equation (\ref{refining_coarse}).

\begin{center}
\begin{equation}\label{refining_coarse}
\scalebox{0.87}{%
$\begin{aligned}
    &duplicate(S,r)=\left\{\overbrace{s_{i},...,s_{i}}^{\textit{r times}}\right\}_{i = 1}^{N} \\
    &Q' = duplicate(S,r) \oplus \mathit{S'^{\Delta}}
\end{aligned}$%
}
\end{equation}
\end{center}

\subsection{Self-Attention Block} 
\label{sec:MHSA} 
\quad Inspired by self-attention, which learns long-range dependency \cite{vaswani2017attention}, we use MultiHead Self-Attention (MSA) was used to extract the GFs from $Q'$. As shown in Figure \ref{fig:model}, the shape of $Q'$ was changed from $\mathbb{R}^{rN \times 3}$ to $\mathbb{R}^{N \times 3r}$, and the coordinates of $Q'$ were used as features of the original point set $S$. The GFs$\in \mathbb{R}^{N \times K'C'}$ was then extracted using the changed shape $Q'$ as the input to the MSA.

\begin{figure*}[htb!]
  \centering
  \subfloat[Input]{\includegraphics[height=8.5cm]{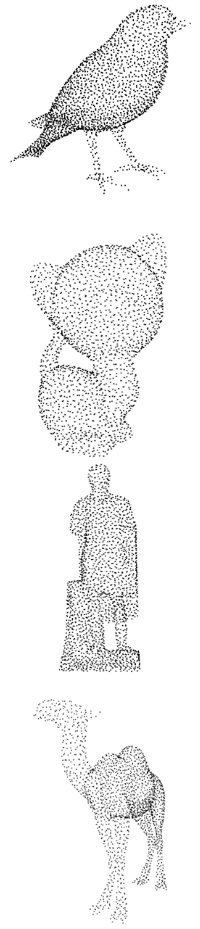}} 
  \subfloat[Dis-PU]{\includegraphics[height=8.5cm]{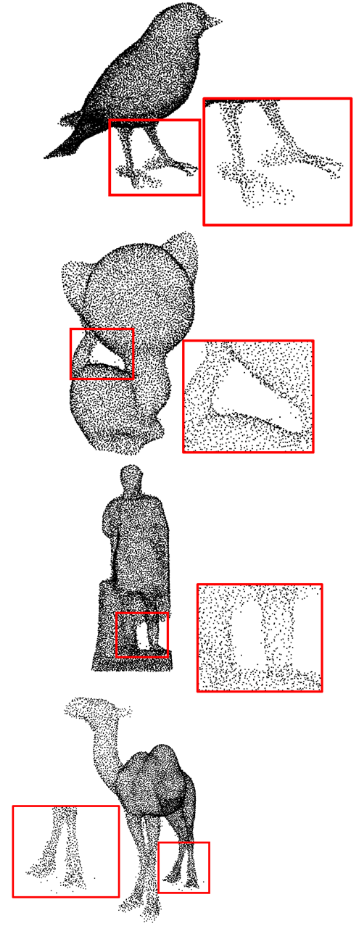}} 
  \subfloat[PU-EVA]{\includegraphics[height=8.5cm]{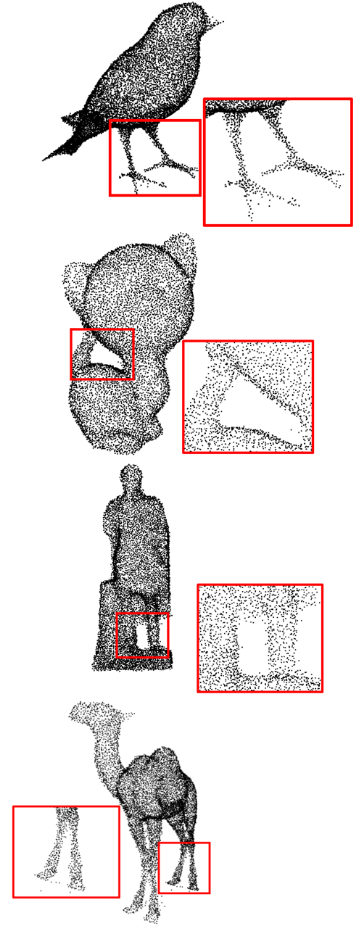}} 
  \subfloat[PU-Transformer]{\includegraphics[height=8.5cm]{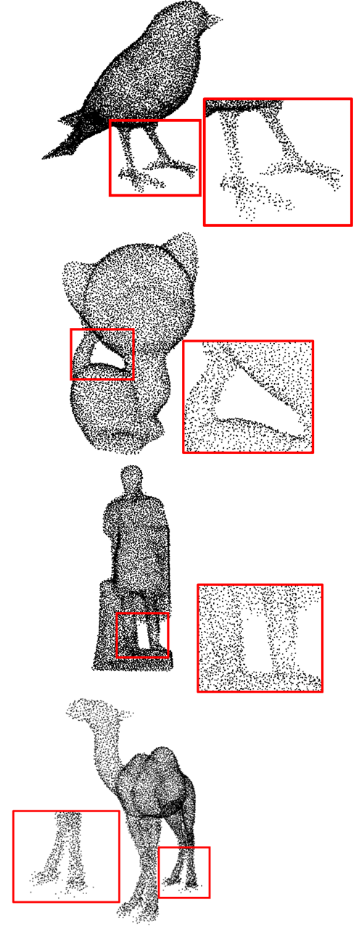}} 
  \subfloat[PU-MFA(Ours)]{\includegraphics[height=8.5cm]{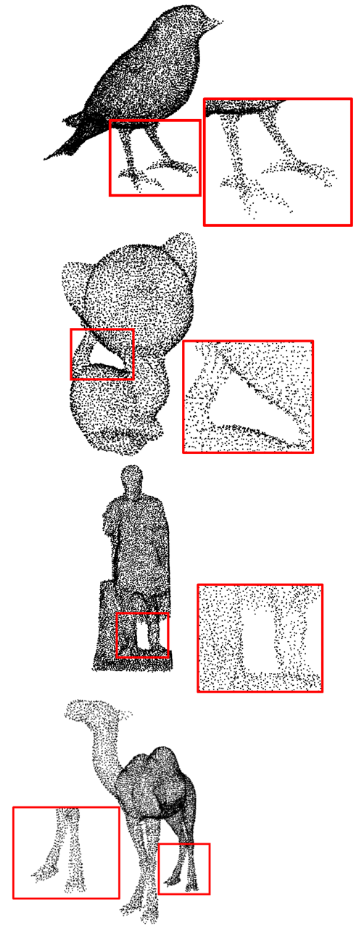}} 
  \subfloat[GT]{\includegraphics[height=8.5cm]{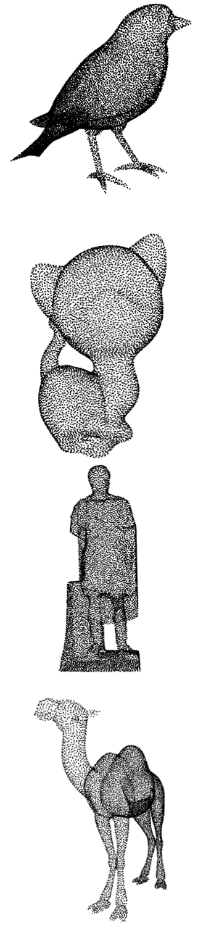}}
  \caption{Visualization result of $\times 4$ up-sampling on PU-GAN dataset.}
  \label{fig:x4_quality}
\end{figure*}

\section{Experimental Settings}
\label{sec:experiment_settings}

\subsection{Datasets}
\label{sec:dataset}
 \quad All methods were trained using the most popular PU-GAN \cite{li2019pu} dataset in these experiments and evaluated using the PU-GAN dataset and the KITTI \cite{geiger2013vision} dataset. The PU-GAN dataset was a synthetic point cloud dataset produced from 147 3D meshes, and the KITTI dataset was a real-scanned point cloud dataset collected using real LiDAR.
 
\quad The training phase used 120 3D meshes from the PU-GAN dataset. All patches were generated via the Poisson disk sampling after converting the original mesh to a point cloud, just like the patch-based up-sampling approach. The sampling resulted in 24,000 input-output pairs.

 \quad In the evaluation phase, 27 3D meshes from the PU-GAN dataset were converted into point clouds to test the synthetic point up-sampling, and the real-scanned point up-sampling test was performed using the KITTI dataset. The generated patches should cover all point sets when evaluating the synthetic point cloud and real-scanned point cloud up-sampling. After merging each up-sampled patch, the up-sampled point set was reconstructed by farthest point sampling. More details can be found at study in PU-GAN \cite{li2019pu}. This dataset was downloaded and used from \url{https://github.com/liruihui/PU-GAN}.

\subsection{Loss Function} 
 \quad In most point cloud reconstruction methods, CD is used as the loss function \cite{nguyen2019graphx,wen2020point,yu2021pointr}. However, it was confirmed empirically that the Density-Aware Charmfer Distance showed good performance, considering the uniformity of the points set on the CD. Therefore, the total loss was formulated as equation (\ref{loss_func}), where $\alpha$ is linearly interpolated from 0.1 to 1 during training and $\left\| \cdot\right\|_{2}$ is $L_{2}$ norm.
 
\begin{center}
\begin{equation}\label{loss_func}
\scalebox{0.75}{%
$\begin{aligned}
    &\mathit{Loss(Q',Q,D)} = \mathit{L_{CD}(Q',D)} + \alpha *\mathit{L_{DCD}(Q,D)} \\ 
    &\mathit{L_{CD}(Q',D)} = \frac{1}{\left|Q'\right|}\sum_{x\in Q'} \underset{{y\in D}}{\mathrm min}\left\|x-y\right\|_2 + \frac{1}{\left|D\right|}\sum_{y\in D} \underset{x\in Q'}{min}\left\|y-x\right\|_2  \\
    &\mathit{L_{DCD}(Q,D)} = \frac{1}{\left|Q\right|}\sum_{x\in Q}^{}\underset{y\in D}{min}(1 - e^{-\left\|x-y\right\|_2})+\frac{1}{\left|D\right|}\sum_{y\in D}^{}\underset{x\in Q}{min}(1 - e^{-\left\|y-x\right\|_2})
\end{aligned}$%
}
\end{equation}
\end{center}

\begin{table*}[htb!]
\centering
\resizebox{\textwidth}{!}
{%
\begin{tabular}{@{}c|cccc|cccc@{}}
\toprule
\multirow{2}{*}{\textbf{Method}} & \multicolumn{4}{c|}{$\boldsymbol{\times}$\textbf{4}(\textbf{2048} $\boldsymbol{\rightarrow}$ \textbf{8192})}                                       & \multicolumn{4}{c}{$\boldsymbol{\times}$\textbf{16}(\textbf{512} $\boldsymbol{\rightarrow}$ \textbf{8192})}                              \\ \cmidrule(l){2-9} 
                        & \textbf{CD}$\mathbf{(10^{-3})}$     & \textbf{HD}$\mathbf{(10^{-3})}$   & \textbf{P2F}$\mathbf{(10^{-3})}$   & \textbf{\#Params(M)} & \textbf{CD}$\mathbf{(10^{-3})}$ & \textbf{HD}$\mathbf{(10^{-3})}$ & \textbf{P2F}$\mathbf{(10^{-3})}$ & \textbf{\#Params(M)} \\ \midrule
\textbf{Dis-PU}         & 0.2703          & 5.501          & 4.346          & \textbf{2.115}     & 1.341           & 28.47           & 20.68            & \textbf{2.115}                  \\
\textbf{PU-EVA}         & 0.2969           & 4.839              & 5.103              & 2.198              & 0.8662           & 14.54           & 15.54            & 2.198                  \\
\textbf{PU-Transformer} & 0.2671          & 3.112          & 4.202          & 2.202              & 1.034           & 21.61           & 17.56            & 2.202                  \\ \midrule
\textbf{PU-MFA(Ours)}           & \textbf{0.2326} & \textbf{1.094} & \textbf{2.545} & 2.172              & \textbf{0.5010}           & \textbf{5.414}           & \textbf{9.111}            & 2.172                  \\ \bottomrule
\end{tabular}}
\caption{Comparing the quantitative evaluation of $\times 4$ and $\times 16$ up-sampling with the state-of-the-art methods.}
\label{tab:pugan_test}
\end{table*}

\subsection{Metric}
\quad This study evaluated the method using CD, HD, and P2F metrics, as in previous studies \cite{li2021point,luo2021pu,qiu2021pu}. CD is a metric that measures the similarity between a set of GT  points and a set of predicted points for each point, and HD is an evaluation metric that measures the outliers in a set of predicted points based on a set of GT points. P2F is an index that measures the similarity between the original mesh and the predicted point set and measures the quality of the predicted point set. The parameter complexity was also measured by measuring the number of parameters. For all metrics, a lower the number, meant better performance.

\subsection{Comparison Methods}
\quad The proposed method was compared with three state-of-the-art methods: Dis-pu \cite{li2021point}, PU-EVA \cite{luo2021pu}, and PU-Transformer \cite{qiu2021pu} to validate the method. For an exact comparison, all methods were implemented using pytorch \cite{NEURIPS2019_9015} version 1.7.0 on Ubuntu 20.04 and trained on the same Intel i9-10980XE CPU and NVIDIA TITAN RTX environment.

\begin{figure}[htb!]
  \centering
  \subfloat[Input]{\includegraphics[width=0.15\columnwidth]{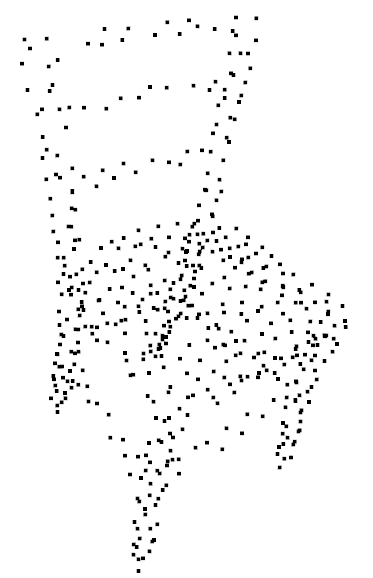}} 
  \subfloat[Dis-PU]{\includegraphics[width=0.15\columnwidth]{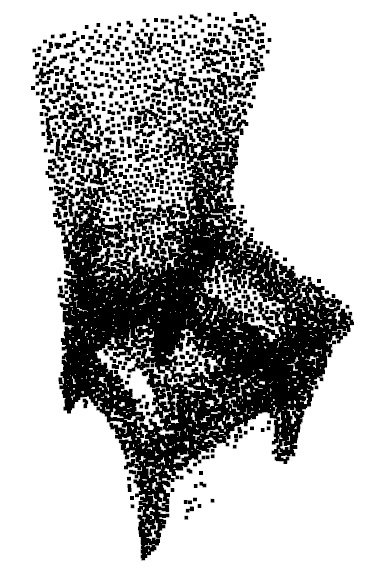}} 
  \subfloat[PU-EVA]{\includegraphics[width=0.15\columnwidth]{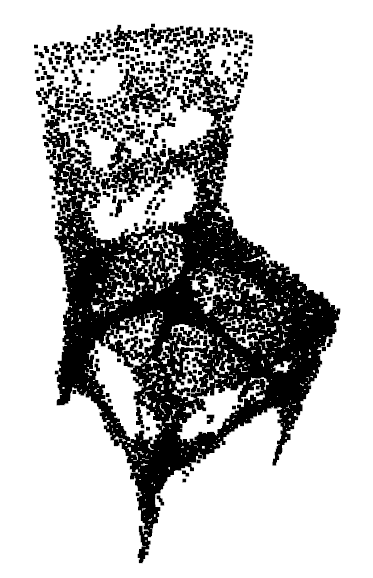}}
  \subfloat[PU-Transformer]{\includegraphics[width=0.15\columnwidth]{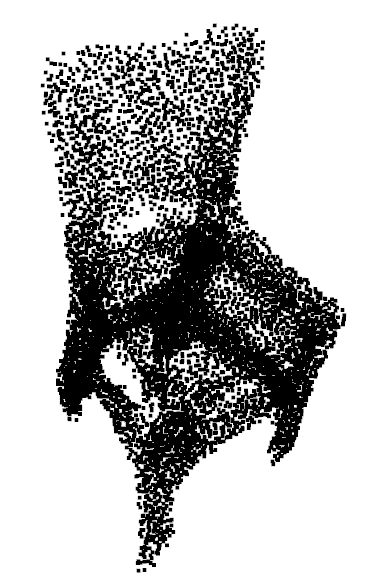}} 
  \subfloat[PU-MFA(Ours)]{\includegraphics[width=0.15\columnwidth]{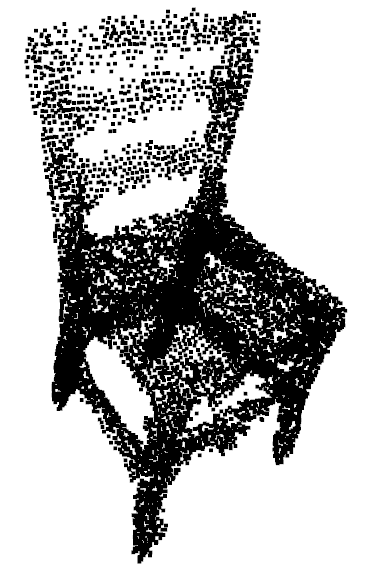}} 
  \subfloat[GT]{\includegraphics[width=0.15\columnwidth]{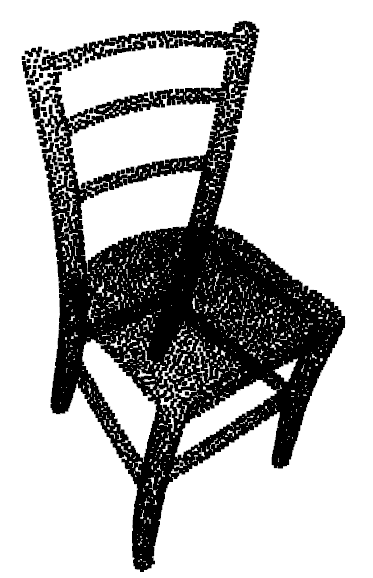}}
  \caption{Visualization result of $\times16$ up-sampling of PU-GAN dataset.}
    \label{fig:x16}
\end{figure}

\subsection{Implementation Details}
\label{sec:implementation}
\quad All methods for the experiment were trained with a batch size of 64 for 100 epochs, and the Adam \cite{kingma2014adam}  optimizer with a learning rate of 0.0001 was used. The patch size of KNN used in PT is set to 20 as in PU-Transformer \cite{qiu2021pu}. Rotation, scaling, random perturbation, and regularization were applied to the training dataset. as in previous studies \cite{yu2018pu,li2019pu}. The up-sampling ratio $r$ was four and the input patch size $N$ was 256. The CPG's $C'$ and $K'$ were 32 and 8, respectively. For MFE and GCR, $C$ and $K$ were 16 and 4, respectively. The layer depth of MFE and GCR, $H$, was four. The head number of MCA and MSA was set to eight, as in the previous study \cite{vaswani2017attention}.

\section{Experimental Results}
\label{sec:experiment_results}
\quad Dis-PU \cite{li2021point}, PU-EVA \cite{luo2021pu}, and PU-Transformer \cite{qiu2021pu}, and the present method were compared using the PU-GAN \cite{li2019pu} and the KITTI \cite{geiger2013vision} datasets.

\subsection{Results on 3D Synthetic Datasets}
\quad Table \ref{tab:pugan_test} lists the quantitative performance comparisons for $\times 4$ and $\times 16$ up-sampling. $\times 4$ up-sampling sampled 2,048 points to 8,192 points. $\times 16$ up-sampling sampled 512 points to 8,192 points by repeating the $\times 4$ up-sampling twice. As shown in Table \ref{tab:pugan_test}, the present method showed good performance compared to the other state-of-the-art methods. Although it is not the most efficient in parameter complexity, it showed good efficiency compared to the performance of the present method. 

\quad Figure \ref{fig:x4_quality} presents the visualization result of $\times 4$ up-sampling, and Figure \ref{fig:x16}  is the visualization result of $\times 16$ up-sampling. Figures \ref{fig:x4_quality} (b), (c), and (d), show the set of points representing the bird's leg, the space between the kitten's body and tail, the statue's leg, and the camel's hoof with noisy or unclear boundaries. However, Figure \ref{fig:x4_quality} (e) shows low-noise and clear boundaries. In Figure \ref{fig:x16} (b) and (d), the chair back does not represent the original shape well, and (c) maintains the shape to some extent, but there is considerable noise. On the other hand, Figure \ref{fig:x16} (e) has relatively little noise and represents the original shape well.

\subsection{Results on Real-scanned Datasets}
\begin{figure}[htb!]
  \centering
  \subfloat[Input]{\includegraphics[height=3.cm]{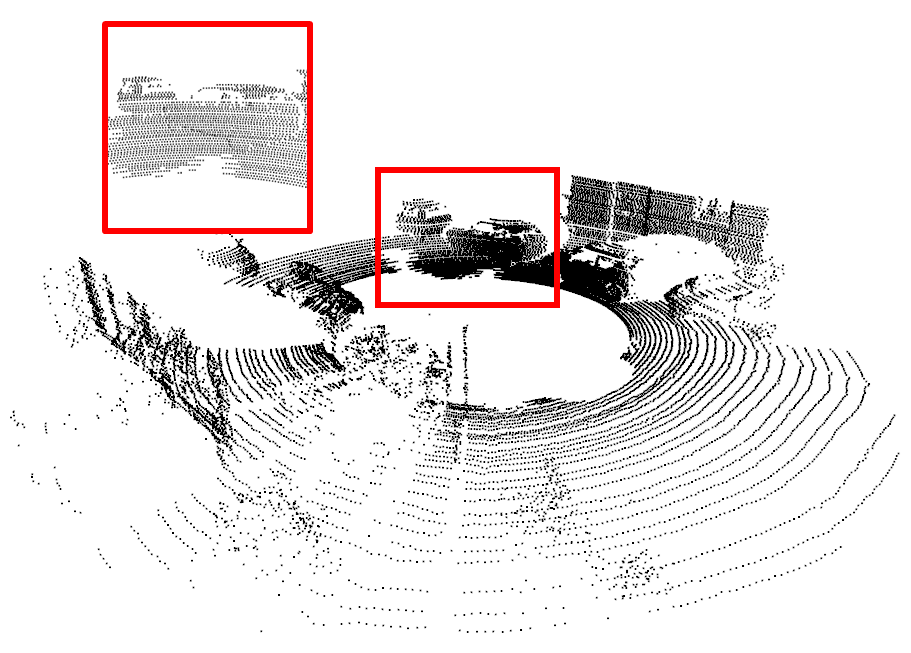}} 
  \subfloat[Dis-PU]{\includegraphics[height=3.cm]{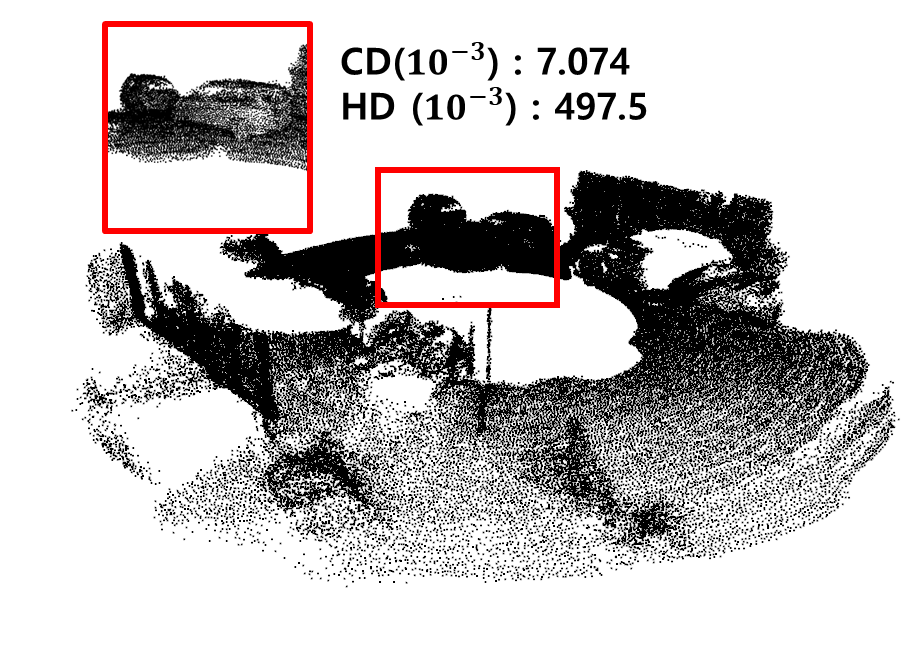}}
  \subfloat[PU-EVA]{\includegraphics[height=3.cm]{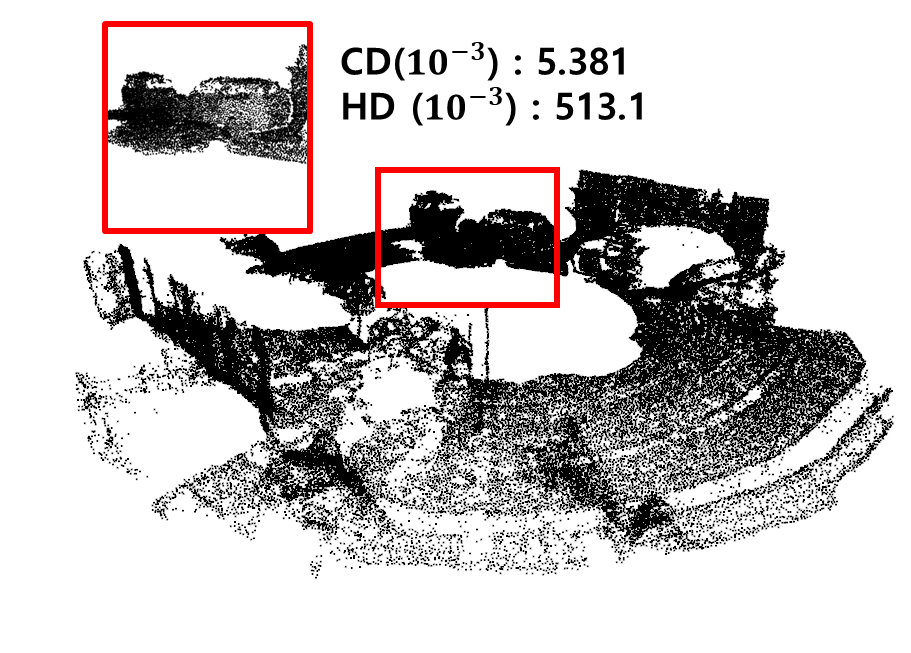}}
  
  \subfloat[PU-Transformer]{\includegraphics[height=3.0cm]{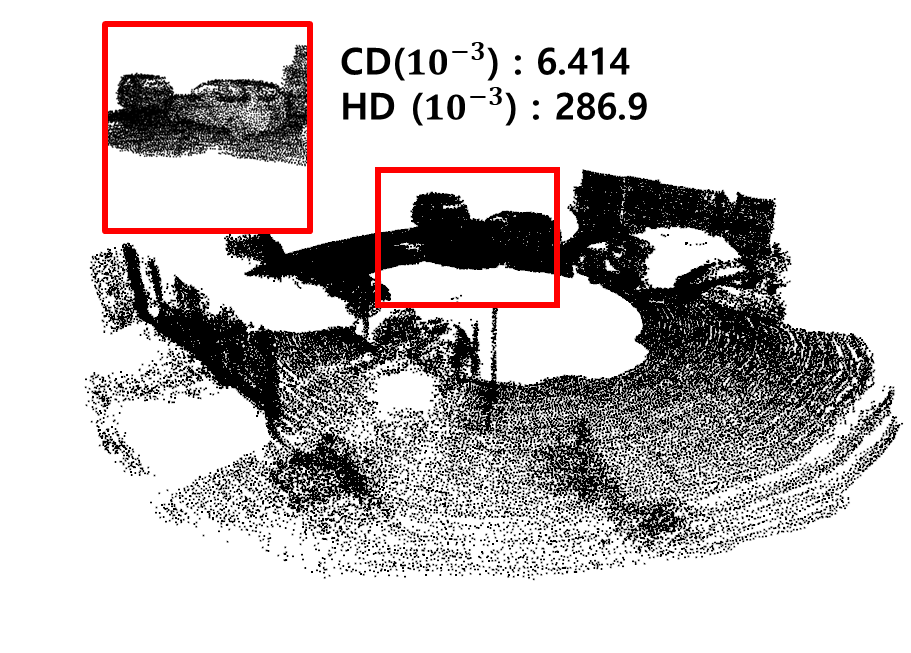}}
  \subfloat[PU-MFA(Ours)]{\includegraphics[height=3.0cm]{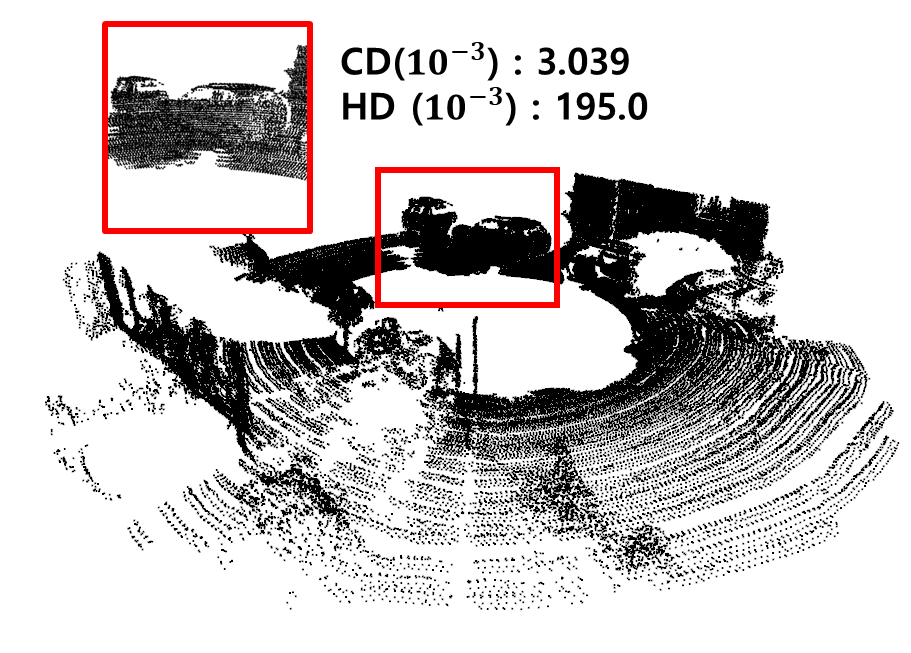}} 
  \subfloat[GT]{\includegraphics[height=3.0cm]{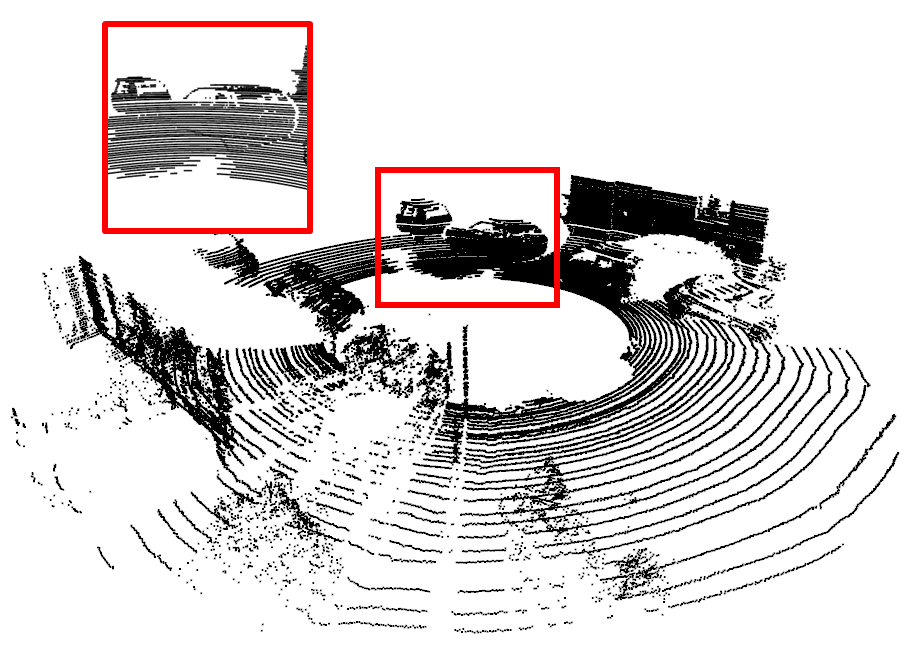}} 
  \caption{Visualization result of $\times 4$ up-sampling of the KITTI dataset.}
    \label{fig:kitti}
\end{figure}

\quad Dis-PU, PU-EVA, PU-Transformer, and the present method were evaluated using the KITTI dataset for $\times 4$ up-sampling. Figure \ref{fig:kitti} shows $\times 4$ up-sampling. In Figure \ref{fig:kitti} (b), (c), and (d), the boundary between the window and the door of the vehicle was unclear. However, Figure \ref{fig:kitti} (e) generated by the present method, showed that the boundary was clearer.

\subsection{Ablation Study}
\quad This method, was evaluated by performing various ablation studies using the PU-GAN dataset.

\subsubsection{Effect of Components}
\quad To demonstrate the effectiveness of the contribution, four cases were divided into ablation studies. The cases were as follows. : Case 1 was a structure using GCR, CPG, and SAB, with the MultiHead Attention (MHA) of GCR and SAB consisting of self-attention with one head. Case 2 was a structure changed from Case 1 to eight heads. Case 3 was a structure using GCRA composed of MCA by adding MFE to Case 2, where the query of all GCRA becomes $F_4$, the final output of MFE. Case 4 was PU-MFA. As shown in Table \ref{tab:ablation_components}, all contributions affected the method performance.

\begin{table}[htb!]
\centering
\resizebox{0.5\columnwidth}{!}
{%
\begin{tabular}{@{}c|ccc|ccc@{}}
\toprule
\multirow{2}{*}{\textbf{Case}} & \multicolumn{3}{c|}{\textbf{Contribution}}  & \multicolumn{3}{c}{\textbf{Metric}}               \\ \cmidrule(l){2-7} 
                               & \textbf{MHA} & \textbf{MFE} & \textbf{MFs} & \textbf{CD$\mathbf{(10^{-3})}$}     & \textbf{HD$\mathbf{(10^{-3})}$}    & \textbf{P2F$\mathbf{(10^{-3})}$}   \\ \midrule
\textbf{1}                     &              &              &               & 0.3349          & 4.461          & 4.926          \\
\textbf{2}                     &$\surd$              &              &               & 0.2473          & 1.101          & 2.829          \\
\textbf{3}                     &$\surd$              &$\surd$              &               & 0.2500          & 2.735          & 2.737          \\ \midrule
\textbf{4}                     &$\surd$              &$\surd$              &$\surd$               & \textbf{0.2362} & \textbf{1.094} & \textbf{2.545} \\ \bottomrule
\end{tabular}}
\caption{Ablation study results to analyze the effect of the present contribution.}
\label{tab:ablation_components}
\end{table}

\subsubsection{Multi-scale Features Attention Analysis}
\begin{figure}[htb!]
  \centering
  \subfloat[Using the MFs GCRA attention map]{\includegraphics[width=0.7\columnwidth]{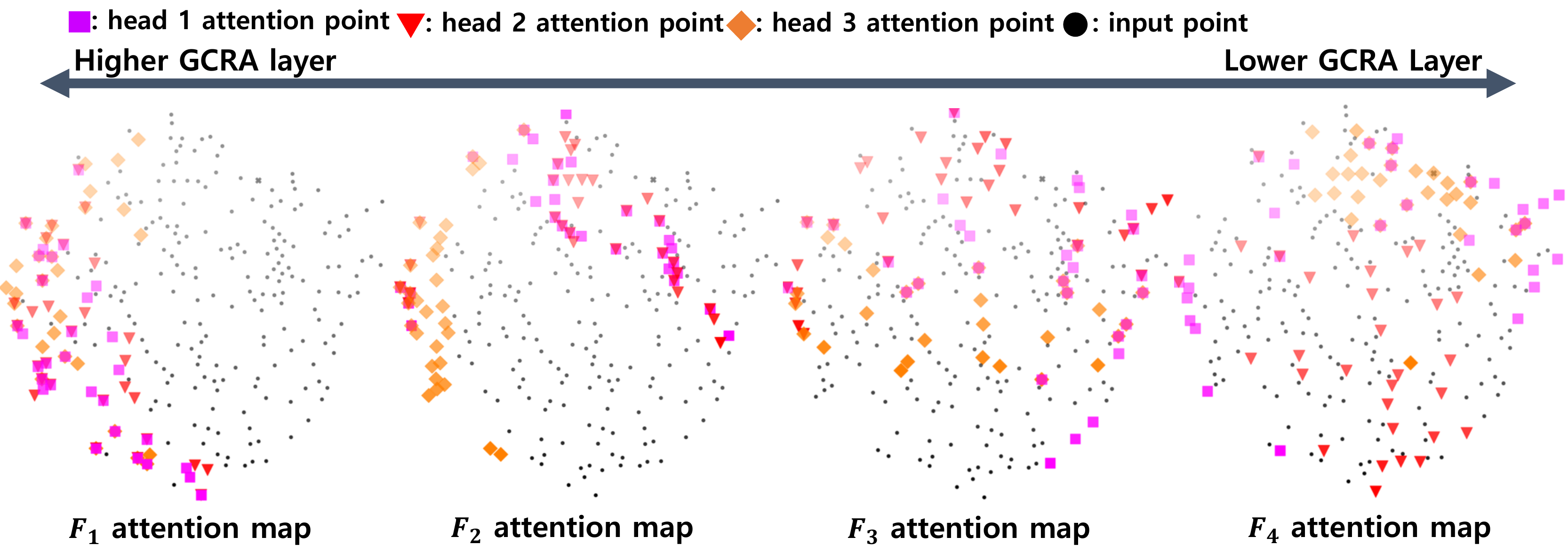}} 
  
  \subfloat[Not Using the MFs GCRA attention map]{\includegraphics[width=0.7\columnwidth]{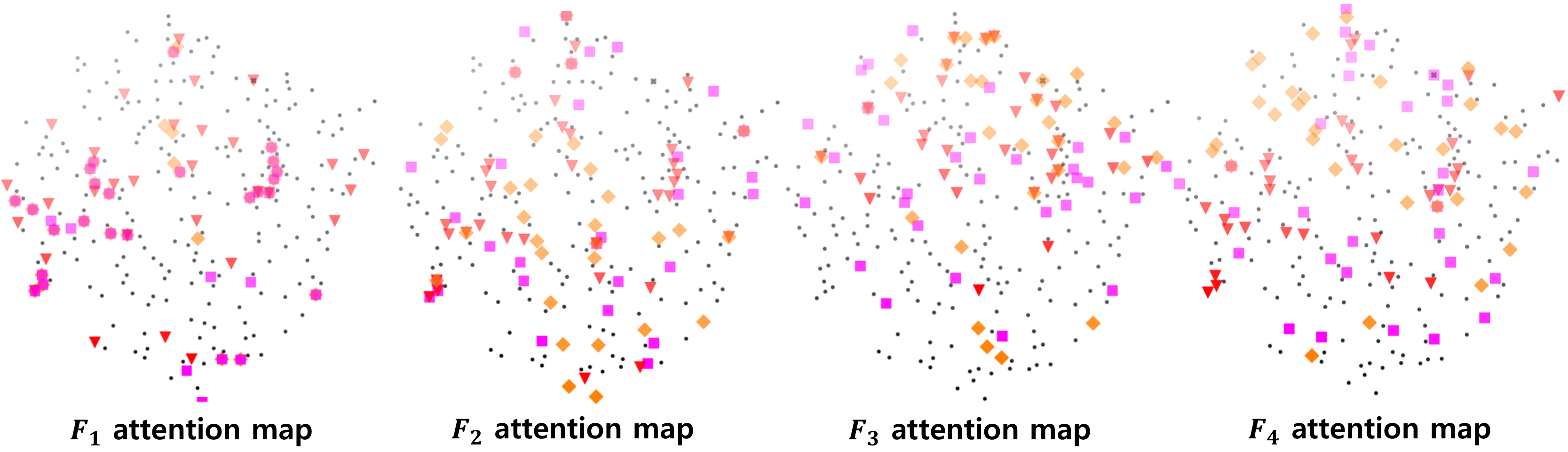}} 
  \caption{Visualization of attention map generated using MFs as a query in GCR with $H=4$.}
    \label{fig:attention_map}
\end{figure}

\quad By visualizing the attention maps of all GCRAs, it was confirmed that the GCRAs of GCR with $H=4$ refined the GFs by adaptively using the MFs extracted from receptive fields of various sizes. Figure \ref{fig:attention_map} shows the results visualized by choosing three attention heads in the GCRA and selecting 30 points, which had the highest attention score in $S$, from each head. The attention map was visualized using Case 3 in Table \ref{tab:ablation_components} without MFs in Figure \ref{fig:attention_map} (b) to compare that MFs operated adaptively. As shown in Figure \ref{fig:attention_map} (a), in the low-layer GCRA, an attention map was formed for a wide range of points in a point set, and in high-layer GCRA, an attention map was formed for a relatively narrow range of points. On the other hand, in Figure \ref{fig:attention_map} (b), a wide range of attention maps was formed regardless of the high and low levels of the hierarchy. This phenomenon confirmed that PU-MFA uses the adaptive point feature for each layer of the GCRA.

\begin{figure}[htb!]
  \centering
\includegraphics[width=0.7\columnwidth]{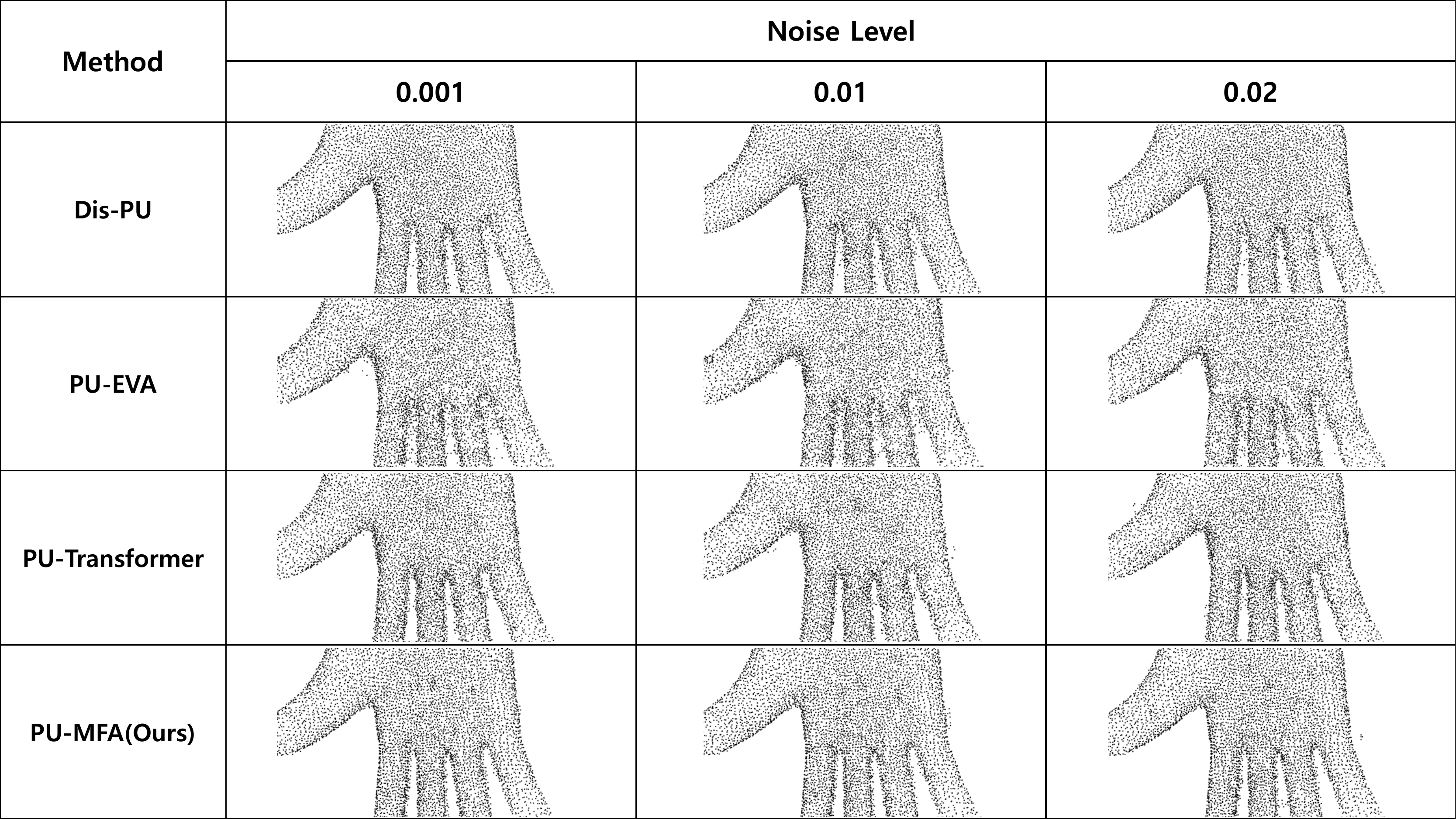} 
  \caption{Visualization result of the effect of noise.}
    \label{fig:noisy_quality}
\end{figure}

\subsubsection{Effect of Noise}
\quad Table \ref{tab:noise_test} lists the $\times 4$ up-sampling results of Dis-PU \cite{li2021point}, PU-EVA \cite{luo2021pu}, PU-Transformer \cite{qiu2021pu}, and the present method using the PU-GAN dataset with various noises added. The noise effect evaluated the result obtained by adding different levels of Gaussian noise $\mathcal{N}$(0,noise level) to a set of input points. As shown in Table \ref{tab:noise_test}, the proposed method showed the most robustness to various noise levels. As shown in Figure \ref{fig:noisy_quality}, it can be seen that the boundary between the fingers blurred in the dense set of points generated by the state-of-the-art methods as the noise level was increased. On the other hand, the proposed method showed that the boundary between the fingers was maintained in the dense set of points generated by the present method.

\begin{table}[htb!]
\centering
\resizebox{0.5\columnwidth}{!}{%
\begin{tabular}{@{}c|cccccc@{}}
\toprule
\multirow{2}{*}{\textbf{Method}} & \multicolumn{6}{c}{\textbf{Various noise levels test at x4 Up-sampling(CD with $\mathbf{10^{-3}}$)}}                                                            \\ \cmidrule(l){2-7} 
                                 & \textbf{0} & \textbf{0.001} & \textbf{0.005} & \textbf{0.01} & \textbf{0.015} & \textbf{0.02} \\ \midrule
\textbf{Dis-PU}                  & 0.2703     & 0.2751         & 0.2975         & 0.3257        & 0.3466         & 0.3706        \\
\textbf{PU-EVA}                  & 0.2969     & 0.2991         & 0.3084         & 0.3167        & 0.3203         & 0.3268        \\
\textbf{PU-Transformer}          & 0.2671     & 0.2717         & 0.2905         & 0.3134        & 0.3331         & 0.3585        \\ \midrule
\textbf{PU-MFA(Ours)}                    & \textbf{0.2326}     & \textbf{0.2376}         & \textbf{0.2547}         & \textbf{0.2764}        & \textbf{0.2989}         & \textbf{0.3195}        \\ \bottomrule
\end{tabular}%
}
\caption{Quantitative evaluation results of the noise effects using the PU-GAN dataset.}
\label{tab:noise_test}
\end{table}

\section{Conclusion}
\label{sec:conclusion}
\quad In this paper, we proposed PU-MFA, a point cloud up-sampling method of U-Net structure that combines multi-scale features and attention mechanism. One of the most significant differences from the previous point cloud up-sampling methods was that PU-MFA used multi-scale features adaptively and effectively through fusion with the cross-attention mechanism. Also, the PU-MFA is the first method to apply the cross-attention mechanism to point cloud up-sampling to the best of the authors’ knowledge. Various experiments were performed on PU-MFA and other state-of-the-art methods using the PU-GAN and the KITTI dataset. As a result, PU-MFA showed better performance than other state-of-the-art methods in various experiments. In addition, ablation study showed that multi-scale features are very useful in PU-MFA for generating high-quality point sets by choosing receptive field size adaptively for each layer.

\quad Despite the successful performance of PU-MFA, PU-MFA cannot cope with an arbitrary up-sampling ratio. A method that can respond to an arbitrary up-sampling ratio is planned in the future to overcome this limitation.

\section{Acknowledgements}
\quad This research was supported by BK21 Program(5199990814084) through the National Research Foundation of Korea (NRF) funded by the Ministry of Education and Korea Institute of Police Technology (KIPoT) grant funded by the Korea government (KNPA) (No.092021C26S03000, Development of infrastructure information integration and management technologies for real time traffic safety facility operation).

\bibliographystyle{unsrt}  
\bibliography{references}

\end{document}